\documentclass[conference,a4paper]{IEEEtran}

\usepackage{booktabs}
\usepackage{cite}
\usepackage{graphicx}
\usepackage{multirow}
\usepackage{stfloats}
\usepackage{tabularx}
\usepackage{tikz}
\usepackage{url}
\usepackage{xcolor}
\usepackage{enumitem}
\usepackage{algorithm}
\usepackage[noend]{algpseudocode}
\usepackage{hyperref}
\usepackage[subtle,margins=normal,leading=normal]{savetrees} 

\definecolor{nvgreen}{HTML}{76B900}

\newcommand{\autofl}{Auto-FL-Research}
\newcommand{\autoflshort}{AFR}
\newcommand{\nvflare}{NVFlare}
\newcommand{\flamby}{FLamby}
\newcommand{\leaf}{LEAF}
\newcommand{\scoregain}[1]{\textbf{#1}}
\newcolumntype{Y}{>{\raggedright\arraybackslash}X}
\graphicspath{{figures/}}

\begin{document}

\title{Auto-FL-Research: Agentic Search for Federated Learning Algorithms}

\author{%
  \IEEEauthorblockN{Holger R. Roth,
    Ziyue Xu,
    Chester Chen,
    Daguang Xu,
    Peter Cnudde,
    Andrew Feng}%
  \textit{NVIDIA, Santa Clara, USA}
}


\maketitle

\begin{abstract}
Federated learning (FL) research often depends on many small but consequential algorithmic choices:
optimizer variants, server aggregation rules, local training schedules, normalization, regularization,
and model architecture. These choices are expensive to explore manually and difficult to compare fairly when candidate changes can also alter the FL training or evaluation path. In this work, we present \autofl{} (\autoflshort{}), a constrained coding-agent workflow for FL algorithmic
recipe search. Agents may propose and implement candidate training algorithms, including server aggregation rules,
client update schedules, local objectives, and registered model variants, while task profiles fix the
mutation surface, compute budget, communication contract, and final model evaluation.
Each campaign records candidate scores, runtime, edited files, artifacts, and failure status.

We evaluate \autoflshort{} on five healthcare cross-silo \flamby{} tasks and on grouped-client profiles for the five fixed \leaf{} datasets plus the \leaf{} synthetic task. Five-seed repeat evaluations support gains on four \flamby{} tasks and five of six \leaf{} profiles, while also exposing seed-sensitive and search-selected failure cases. Same-budget controls show that several gains correspond to FL-recipe changes, whereas other improvements are recovered by fixed-surface scalar controls or fail under repeat or held-out evaluation.
These mixed outcomes are part of the contribution: they show how agent-generated candidates can be separated into repeated FL mechanisms, fixed-surface tuning effects, and selected single-run artifacts.
\end{abstract}


\section{Introduction}

Federated learning (FL) promises collaborative model development without centralizing raw data, but the practical performance of an FL system depends on a large design surface~\cite{mcmahan2017fedavg}. A practitioner must choose local optimizers, server aggregation rules, schedules, regularization, client participation, model architecture, evaluation strategy, and many task-specific details~\cite{kairouz2021advances}. These choices interact with data heterogeneity and communication constraints, so improvements that appear obvious in centralized training can fail in FL.

Automated FL methods have explored specific portions of this surface, including learnable aggregation, federated hyperparameter optimization, federated neural architecture search, and adaptive server optimizers \cite{yang2021autofedavg,xu2021fednas,guo2022autofedrl,reddi2021fedopt}. However, many useful research advances are not a single scalar hyperparameter. A competitive FL algorithm may require introducing a new model architecture, changing a local loss, adding a server optimizer, or using an improved server aggregation method while preserving the protocol and the benchmark definition.

\begin{figure}[htbp]
  \centering
  \includegraphics[width=0.95\columnwidth]{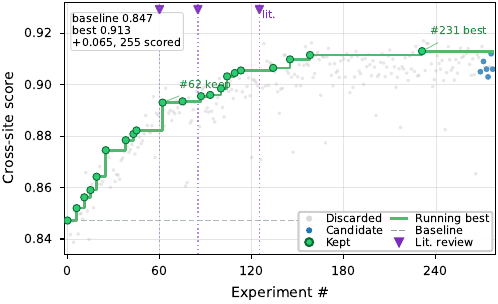}
  \caption{Illustrative CIFAR-10 Auto-FL-Research campaign progress. Each point is a candidate in the run log; gray points are discarded candidates, blue points are active candidates, green points are kept candidates, and the green step line tracks the running best final global-model score. Purple markers indicate logged literature-review events.}
  \label{fig:progress}
\end{figure}

Recent coding agents make it possible to automate code-level research loops, but unconstrained
experimentation can confound evaluation: an agent can change the metric, alter the data split,
silently increase compute, or break the FL contract.
\autofl{} addresses this by fixing what the agent may edit and how every candidate is evaluated.
The agent is instructed and validated to modify code only inside a task-defined mutation surface and must
evaluate candidates through a fixed FL harness, here implemented with NVIDIA FLARE (\nvflare{}).
Each run records its budget, score, status, artifacts, and literature sources.
The implementation corresponding to the methods described in this paper is available as the NVIDIA FLARE Auto-FL research example\footnote{\url{https://github.com/NVIDIA/NVFlare/tree/main/research/auto-fl-research}},
including the control plane, task profiles, plotting utilities, and reporting workflow.

We therefore treat agent changes as candidate-generating steps rather than as final claims.
Candidate scores are interpreted together with the recorded search trace, repeated seed evaluations,
and task controls.
\autoflshort{} is not proposed as a new FL optimizer; it is a constrained research protocol for using coding agents to generate, record, and check candidate FL algorithms under fixed execution and evaluation contracts.

This paper makes three contributions.
\begin{itemize}
  \item We describe a contract-preserving agentic FL search harness based on \nvflare{} task profiles and fixed budgets, and make explicit which code-level mutations are allowed beyond scalar HPO.
  \item We evaluate the harness on \flamby{} healthcare tasks~\cite{duterrail2022flamby} and \leaf{} federated benchmark tasks~\cite{caldas2019leaf}, including five-seed repeats of selected \flamby{} and \leaf{} configurations with matched baselines.
  \item We analyze which agent-discovered FL mechanisms transfer across tasks, identify search-selected gains that do not survive repeat or held-out evaluation, and use same-budget controls to distinguish FL-specific recipe changes from fixed-surface scalar tuning.
\end{itemize}

The intended outcome is therefore not only a better configuration for a benchmark task, but a reproducible
record of what was tried, which ideas transferred, which candidates failed, and which selected gains survived
repeat or held-out evaluation.

\section{Related Work}

\paragraph{Federated Optimization}

Federated Averaging (FedAvg) remains the canonical baseline for cross-device and cross-silo FL \cite{mcmahan2017fedavg}. FedProx adds a proximal term to stabilize optimization in the presence of client heterogeneity \cite{li2020fedprox}. FedOpt generalizes server-side adaptive optimization, including FedAdam-style updates over aggregated client model differences \cite{reddi2021fedopt}. SCAFFOLD uses control variates to reduce client drift \cite{karimireddy2020scaffold}. \autoflshort{} treats these as baseline mechanisms and as building blocks that agents may combine with task-specific local training changes.

\paragraph{Automated FL and Federated Architecture Search}

Prior Automated FL work, building on AutoML and NAS literature~\cite{hutter2019automated,elsken2019neural,he2021automl}, has automated narrower FL design spaces, including federated NAS, learnable aggregation, Bayesian AutoML in FL, client participation, edge-resource scheduling, and FL HPO~\cite{preuveneers2023autofl,kim2021autofl,hu2024autofl,you2023autofl,saadati2024stepwiseautofl,yang2021autofedavg,guo2022autofedrl,xu2021fednas,he2020fednas}.
In contrast, \autoflshort{} is not a single optimizer or controller; it is a constrained coding-agent harness
for code-level FL recipe search under fixed communication and scoring contracts.

\paragraph{Benchmarks and Execution Frameworks}

\flamby{} provides realistic healthcare cross-silo FL tasks with public splits, baseline models, and metrics \cite{duterrail2022flamby}. \leaf{} provides federated datasets for cross-device-style settings, including FEMNIST, Sent140, Shakespeare, CelebA, and Reddit \cite{caldas2019leaf}. The \leaf{} project also distributes a synthetic classification task \cite{leafsite2026}. \nvflare{} provides production-oriented FL execution abstractions and simulation capabilities \cite{roth2022nvflare}. We use \nvflare{} as the execution substrate so that candidate changes are evaluated through an FL runtime rather than a standalone benchmark script.

\paragraph{Agentic Research Loops}

The \autoflshort{} workflow is inspired by emerging autonomous research systems that combine experiment
records, code edits, and literature-guided proposal generation.
EAIRA~\cite{cappello2025eaira} frames the broader problem of evaluating AI models as scientific research
assistants, arguing for assessment beyond static question answering, including controlled lab-style and
field-style evaluations of how models support real research tasks.
End-to-end systems such as The AI Scientist and AI Scientist-v2 automate idea generation, code execution,
experiment analysis, and paper writing for machine-learning research~\cite{lu2024aiscientist,yamada2025aiscientistv2},
while Agent Laboratory studies a more interactive research-assistant workflow with optional human feedback
\cite{schmidgall2025agentlab}.
Karpathy's ``autoresearch'' project demonstrates a minimal agentic loop for repeatedly improving a fixed
training task under a persistent result log~\cite{karpathy2026autoresearch}.
Camyla~\cite{gao2026camyla} emphasizes structured literature search, memory, and proposal generation for
medical image segmentation research.
\autoflshort{} adapts these ideas to federated learning by adding task profiles, communication-contract
invariants, cross-site evaluation, and FL-specific mutation boundaries, so that an agent's contribution is
judged by executable benchmark outcomes rather than by text-only responses or manuscript generation alone.

\section{Method: Agentic Search Harness}

\begin{figure*}[htbp]
\centering
\begin{minipage}[c]{0.62\textwidth}
\centering
\includegraphics[width=\linewidth]{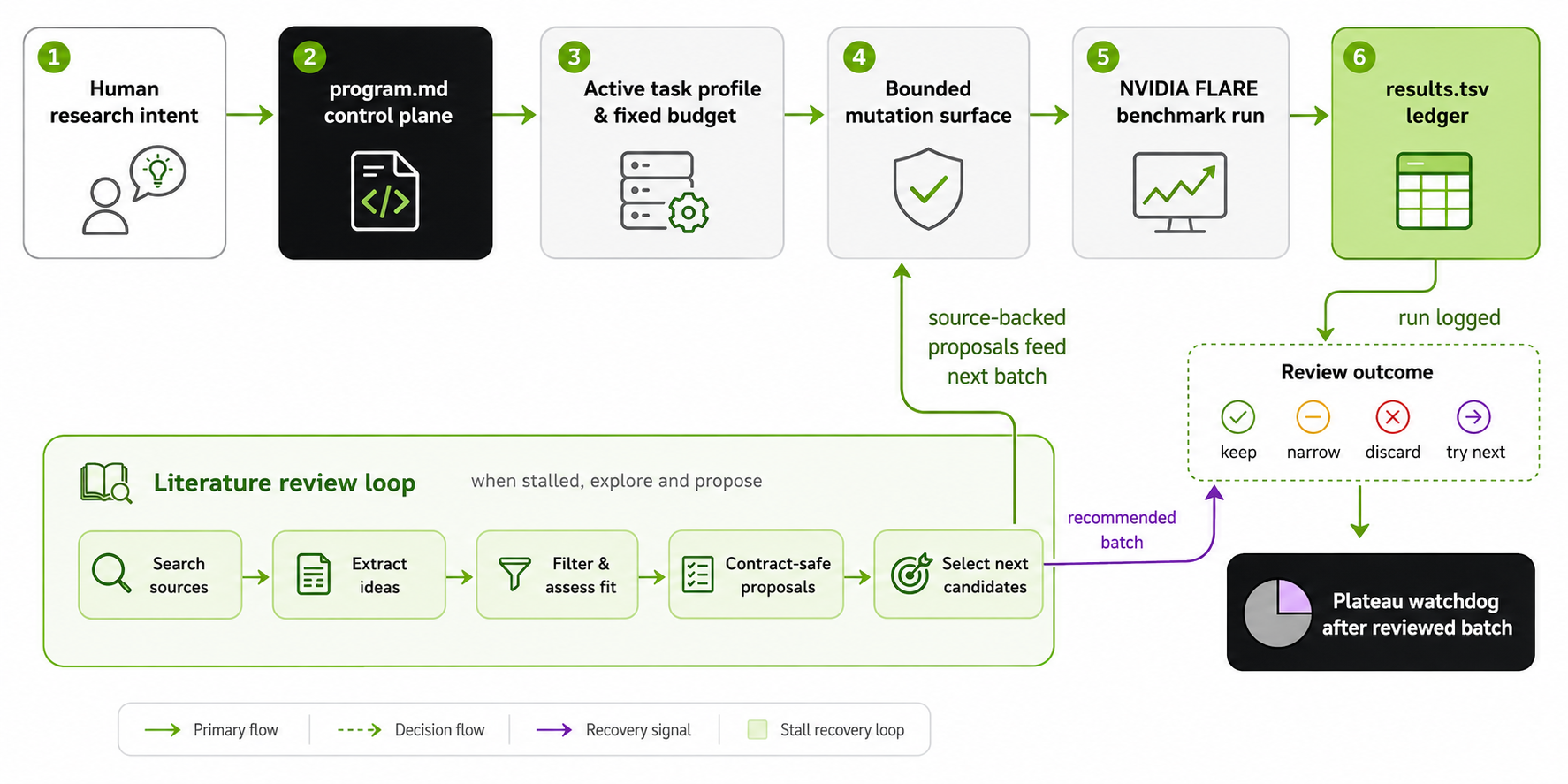}
\end{minipage}\hfill
\begin{minipage}[c]{0.34\textwidth}
\centering
\includegraphics[width=\linewidth]{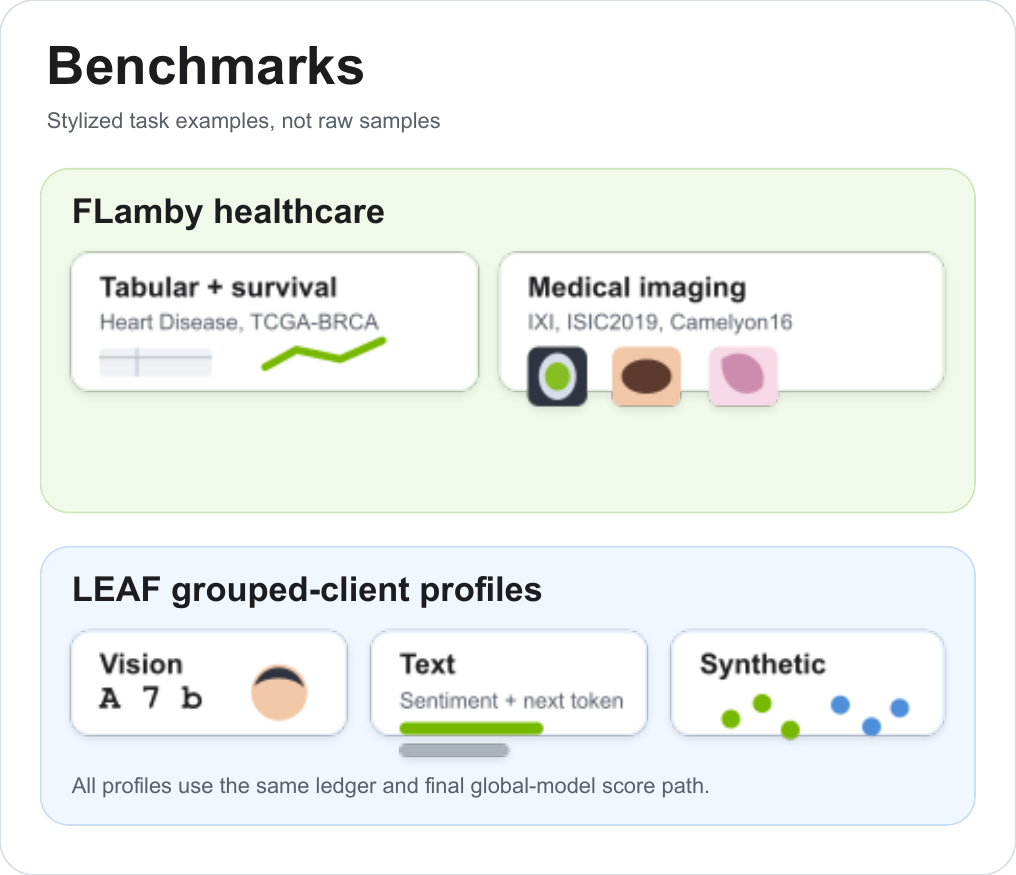}
\end{minipage}
\caption{\autoflshort{} loop and evaluation coverage. \textbf{Left:} the agent starts from research intent, \texttt{program.md}, an active task profile, a fixed budget, and a fixed mutation surface. Candidate \nvflare{} runs append results to \texttt{results.tsv}; reviewed batches are kept, narrowed, discarded, or used to select the next candidate. \textbf{Right:} stylized benchmark modalities from \flamby{} and grouped-client \leaf{} profiles evaluated through the same run log and final global-model scoring path.}
\label{fig:loop}
\end{figure*}

\subsection{Campaign Algorithm}

Algorithm~\ref{alg:campaign} gives the campaign loop used by the agents.
The algorithm is intentionally simple: all state that matters for scientific comparison is either fixed by
the task profile or written into the run record.
The agent may propose code changes, but every candidate must pass the task validation path before it is scored.

\begin{algorithm}[t]
\caption{AFR campaign loop}
\label{alg:campaign}
\footnotesize
\begin{algorithmic}
\Require Task profile, candidate cap, mutation surface, validation commands
\State Initialize \texttt{autoresearch/} branch; run baseline; log to \texttt{results.tsv}
\vspace{0.5em}
\While{budget remains and campaign is not manually stopped}
    \State Execute candidate cycle:
    \begin{enumerate}[leftmargin=1.8em,itemsep=0pt,topsep=1pt]
        \item \textbf{Propose} candidate(s).
        \item \textbf{Validate} edits, budget fields, contract, and smoke test.
        \item \textbf{Run} candidate in \nvflare{}; extract final score.
        \item \textbf{Log} score, runtime, status, description, artifacts.
        \item \textbf{Review} as \emph{keep}, \emph{discard}, or \emph{crash}.
    \end{enumerate}

    \If{plateau watchdog triggers}
        \State \textbf{Recover} via literature loop:
        \Statex \hspace{\algorithmicindent} select source-backed proposals; log event.
    \EndIf
\EndWhile
\vspace{0.5em}
\State \textbf{Finish:} repeat selected configurations; regenerate plots; write final report.
\end{algorithmic}
\end{algorithm}

\subsection{Task Profiles and Fixed Budgets}

Each campaign begins with a task profile that specifies the dataset, metric, model budget, client/site configuration, number of rounds, final evaluation policy, and allowed mutation files. A candidate is comparable only if it preserves the fixed budget fields. For the architecture sub-campaign, the profile includes a maximum parameter count and requires that the selected architecture, normalization mode, and parameter cap be instantiated identically on the server and all clients.

\subsection{Federated Contract}

The agent must preserve the \nvflare{} client contract. In our experiments, clients receive the current global model, load it strictly, perform local training or evaluation, compute a model difference, and send a DIFF-typed update with the number of local steps in metadata. The same final global server model is used for metric evaluation. This prevents a candidate from appearing better by changing the evaluation route, changing the update type, or using a different model state schema on client and server.
In the implementation, this contract is checked by an AST-based static validator that requires
\texttt{flare.init()}, \texttt{flare.receive()}, \texttt{flare.send()}, strict \texttt{state\_dict} loading,
typed update outputs, \texttt{NUM\_STEPS\_CURRENT\_ROUND} metadata, and the evaluation branch.
Each task profile also runs Python compile checks and a task-specific smoke command before full campaign use.

\subsection{Mutation Surface}

The allowed mutation surface includes task-local client training logic, task-local job construction, registered model variants, task-local utilities, and shared custom aggregators. Agents may tune optimizers, schedules, regularization, local step counts, server learning rates, momentum, FedProx-like objectives, FedOpt-style server rules, and architecture variants. They may not change raw data bridges or task data semantics unless the human explicitly asks for a protocol or benchmark change. These boundaries are checked partly by code and partly by review: the static validator catches contract breakage, the smoke run catches many runtime protocol errors, and the final report identifies which files were edited for each kept candidate. The current system does not yet provide a complete cryptographic or sandbox-level proof that forbidden files were untouched; we treat that as an engineering target for future hardening.

\subsection{Run Log, Review, and Literature Loop}

Every candidate is recorded in a tab-separated run log with a score, runtime, budget, status, target file,
description, and artifact paths.
Candidate rows are finalized as \emph{keep}, \emph{discard}, or \emph{crash}.
When the search plateaus, the agent must consult related literature, write down the source-backed idea,
and then implement a candidate.
The final report can then distinguish simple parameter tuning from changes to paper-derived methods.
In the reference workflow, a plateau watchdog recommends switching to literature mode after a sustained run
of scored non-crash candidates without a material improvement or a literature reset.
Literature events are recorded as non-scored rows, so search cost and proposal timing remain visible after
the campaign.

\subsection{Artifact Trail}

The output of a campaign is an artifact trail, not only a best score.
The \autoflshort{} harness keeps the control prompt, task profile, mutation schema, candidate table, generated
progress plot, final report, selected code diffs, and follow-up seed evaluations together in the experiment
branch.
This structure lets a reviewer reconstruct the search surface, identify invalid candidates, separate selected
wins from repeated results, and inspect whether a claimed mechanism came from scalar tuning, task-local code,
architecture registration, or literature-backed proposal generation.
The best configuration is one output of this record, not the only object of analysis.

\section{Experimental Design}

\begin{table*}[b]
\centering
\caption{Search-space comparison for interpreting \autoflshort{} gains beyond scalar HPO.}
\label{tab:search_space}
\small
\setlength{\tabcolsep}{5pt}
\renewcommand{\arraystretch}{1.14}
\begin{tabularx}{\textwidth}{@{}p{0.17\textwidth}Yp{0.22\textwidth}@{}}
\toprule
Search mode & Mutation surface and guardrails & Evidence role \\
\midrule
Scripted scalar HPO &
\emph{Allowed:} existing scalar/categorical knobs: optimizer, learning rate, local steps, FedProx, server
optimizer, scheduler, and regularization. \newline
\emph{Disallowed:} new model architectures, new task-local code, and literature-derived method additions. &
Same-budget control for ordinary HPO. \\
\addlinespace[0.25em]
Architecture-open \autoflshort{} &
\emph{Allowed:} scalar knobs plus registered \texttt{model\_arch} variants under
\texttt{max\_model\_params}; task-local training code within profile bounds. \newline
\emph{Disallowed:} data bridge changes, metric changes, protocol changes, and unregistered architecture drift. &
Tests whether code-level recipe search adds value. \\
\addlinespace[0.25em]
Literature-enabled \autoflshort{} &
\emph{Allowed:} architecture-open surface plus source-backed plateau recovery recorded in the run log. \newline
\emph{Disallowed:} uncited benchmark changes, data/metric edits, and extra compute outside the budget. &
Tests research-loop behavior; causal value requires ablation. \\
\addlinespace[0.25em]
Trajectory repeats &
\emph{Allowed:} independent agent sessions under the same task profile, budget, and mutation schema. \newline
\emph{Disallowed:} changing the campaign objective or fixed budget fields. &
Tests whether mechanisms recur across agent runs. \\
\bottomrule
\end{tabularx}
\par\vspace{0.75em}
\begin{minipage}{0.95\textwidth}
\footnotesize
\centering
All modes preserve the same FL communication contract, data bridge, candidate schema, and final
global-model scoring path.
\end{minipage}
\end{table*}

\subsection{Search Spaces and Controls}

The central comparison in this work is not an unconstrained agent versus a weak manual baseline.
The comparison is between search spaces that differ in what they permit while sharing the same communication
contract, final scoring path, and candidate schema.
Table~\ref{tab:search_space} summarizes the mutation surfaces used for the main evidence blocks.
The scripted HPO controls can tune scalar and categorical knobs already exposed by the harness, but cannot
introduce new task-local code or registered architectures.
The architecture-open \autoflshort{} campaigns can add named model variants under a parameter cap, and the
literature loop can propose source-backed methods, but both remain inside the same validation, smoke-test,
and final-global-model scoring gates.
Concretely, the scripted scalar controls were generated with fixed random seeds and fixed model architectures
as summarized in Table~\ref{tab:scripted_hpo_space}.

\begin{table*}[htbp]
\centering
\caption{Scripted scalar HPO search spaces used for same-budget controls. All candidates used fixed model
architectures. 
}
\label{tab:scripted_hpo_space}
\footnotesize
\setlength{\tabcolsep}{4pt}
\renewcommand{\arraystretch}{1.12}
\begin{tabularx}{\textwidth}{@{}p{0.12\textwidth}p{0.11\textwidth}p{0.21\textwidth}p{0.23\textwidth}X@{}}
\toprule
Task & Fixed arch. & Local / optimizer & Regularization & Aggregation / server optimizer \\
\midrule
FEMNIST &
LEAF baseline &
Steps \{50, 75, 100, 125, 150, 200\};
SGD lr \{0.003, 0.005, 0.0075, 0.01, 0.015, 0.02\};
AdamW lr \{0.0003, 0.001, 0.003\} &
Weight decay \{0, $10^{-5}$, $3{\times}10^{-5}$, $10^{-4}$, $3{\times}10^{-4}$\};
FedProx $\mu$ \{0, $10^{-5}$, $3{\times}10^{-5}$, $10^{-4}$, $3{\times}10^{-4}$, $10^{-3}$\};
label smoothing \{0, 0.03, 0.05, 0.10\};
clip \{0, 1, 5\} &
Aggregators \{weighted, FedAvg, FedAvgM, FedAdam\};
FedAvgM server lr \{0.5, 0.75, 1.0, 1.25, 1.5\};
FedAvgM momentum \{0.3, 0.6, 0.75, 0.9\};
FedAdam server lr \{0.0032, 0.01, 0.0316, 0.1, 0.3162\}. \\
\addlinespace[0.35em]
Sent140 &
LEAF baseline &
Same as FEMNIST, with local steps also including 250 &
Same as FEMNIST &
Same as FEMNIST. \\
\addlinespace[0.35em]
Heart Disease &
FLamby baseline &
Steps up to 250;
SGD lr 0.001--0.02;
AdamW lr 0.0003--0.006 &
Same regularization choices as FEMNIST/Sent140 &
Aggregators also included median;
server lr \{0.0032, 0.01, 0.0316, 0.1, 0.3162, 0.75, 1.0\};
FedAvgM momentum \{0.3, 0.6, 0.75, 0.9\};
FedAdam $(\beta_1,\beta_2,\tau)=(0.9,0.999,10^{-8})$. \\
\addlinespace[0.35em]
ISIC2019 &
Pretrained FLamby baseline &
Steps up to 200;
SGD lr 0.002--0.02;
AdamW lr 0.0003--0.003 &
FedProx $\mu$ up to $3{\times}10^{-3}$;
same weight decay, label smoothing, and clipping choices as above &
Aggregators \{weighted, FedAvg, FedAvgM, FedAdam\};
server lr \{0.001, 0.0032, 0.01, 0.0316, 0.1, 0.3162, 0.75, 1.0\};
FedAvgM momentum \{0.3, 0.6, 0.75, 0.9\};
FedAdam $(\beta_1,\beta_2,\tau)=(0.9,0.999,10^{-8})$. \\
\bottomrule
\end{tabularx}
\end{table*}

\subsection{FLamby Campaigns}

We evaluate five \flamby{} tasks: Fed-Heart-Disease, Fed-TCGA-BRCA, Fed-IXI, Fed-ISIC2019, and Fed-Camelyon16. The reported score is always the final \nvflare{} global server model evaluated via the task harness; higher scores are better for all metrics. Campaigns ran on a local node with four NVIDIA H100 80 GB GPUs, but the reported searches launched one candidate at a time, with each candidate occupying one GPU. Each architecture-open campaign was capped at 100 candidates. Candidates were launched sequentially to avoid concurrent edits to shared task files. Unless otherwise noted, campaigns used the same fixed coding-agent backend\footnote{Codex GPT-5.5 with xHigh effort and Auto-Review.}, prompt/control-plane files, task profile, mutation schema, validation commands, timeout policy, and scoring path. Human intervention was limited to campaign setup, interruption, and post-hoc review.

We compare against three values. The first is the \nvflare{} campaign baseline, i.e., the first fixed-budget run in the same harness. The second is the best \autoflshort{} score found in the campaign. The third is an external target selected from \flamby{} or closely related published work: the original \flamby{} reference, FENS one-shot ensembling results \cite{allouah2024fens}, or FedCompass for IXI \cite{li2024fedcompass}. These targets are useful calibration points, but they do not always align with our exact \nvflare{} protocol, compute budget, or final global model scoring. Therefore, we report repeat-seed means for the strongest candidates and distinguish stable contextual gains from seed-sensitive campaign bests.

\subsection{LEAF Campaigns}

We evaluate the five fixed LEAF benchmark datasets from the original paper~\cite{caldas2019leaf}--FEMNIST, Sent140, Shakespeare, CelebA, and Reddit--plus the LEAF synthetic classification task. Our \leaf{} adapter is a grouped-client approximation: each \nvflare{} client represents a deterministic group of original \leaf{} users, rather than a single physical device. This preserves user records and train/test splits while keeping \nvflare{} simulation costs manageable. Consequently, the \leaf{} experiments test whether \autoflshort{} can improve task profiles under a consistent FL harness, rather than whether it exactly reproduces the large-scale cross-device setup of the original paper.

\subsection{Post-Selection Evaluation Protocol}
During the search, the agent observes the same profile score reported in the campaign log.
This is appropriate for studying benchmark optimization behavior, but it can overfit the reported score.
We therefore treat single-seed campaign bests as selected candidates rather than final statistical claims.
For \flamby{} and \leaf{}, selected configurations and corresponding baselines were repeated with five seeds
and saved with individual per-seed scores, mean, standard deviation, standard error, and
confidence-interval summaries.
For repeated comparisons, we report paired mean differences with descriptive uncertainty half-widths computed
from matched-seed winner-minus-baseline differences.
Because $n=5$ is small and the candidates were selected by search, these intervals should not be interpreted
as confirmatory hypothesis tests.
As an additional overfitting check, we ran a validation-selected and held-out-reported procedure for
Heart Disease and FEMNIST: candidate selection used a deterministic validation subset of each site's
evaluation split, while the selected candidate and matched baseline were rerun on the complementary held-out
subset.
This check is not a replacement for externally defined test sets, but it directly tests whether a selected
candidate survives a score it did not observe during search.
We use candidate wall-clock runtime as the normalized experimental cost.
Agent-session token telemetry is captured in post-campaign reports when the agent runtime exposes it, but it
is treated as artifact metadata rather than as the search budget because telemetry availability differs across
agent frontends.


\begin{table}[t]
\centering
\caption{Campaign accounting for the main reported searches. All rows used a 100-candidate cap and logged 100 candidates. Lit. denotes literature events;
Wall-h excludes non-scored literature rows and follow-up repeat evaluations.
}
\label{tab:budget}
\scriptsize
\setlength{\tabcolsep}{3pt}
\begin{tabular}{llrrrr}
\toprule
Suite & Task & Crash & Lit. & Min/cand. & Wall-h \\
\midrule
FLamby & Heart & 1 & 1 & 1.1 & 1.9 \\
FLamby & TCGA & 1 & 1 & 0.9 & 1.5 \\
FLamby & IXI & 3 & 1 & 42.8 & 71.3 \\
FLamby & ISIC & 2 & 2 & 5.4 & 9.1 \\
FLamby & Camelyon & 0 & 3 & 4.5 & 7.4 \\
LEAF & FEMNIST & 1 & 0 & 1.8 & 3.1 \\
LEAF & Shakespeare & 2 & 0 & 7.5 & 12.5 \\
LEAF & Synthetic & 0 & 0 & 1.4 & 2.3 \\
LEAF & Sent140 & 1 & 2 & 2.1 & 3.5 \\
LEAF & CelebA & 0 & 0 & 2.4 & 4.0 \\
LEAF & Reddit & 0 & 0 & 2.5 & 4.2 \\
\bottomrule
\end{tabular}
\end{table}

\section{Results}

\subsection{FLamby Healthcare Tasks}

Table~\ref{tab:flamby} summarizes the five completed \flamby{} campaigns with five-seed repeats of the selected configuration and matched baseline. The largest reproducible gains were observed on IXI, ISIC2019, and Camelyon16 (see Fig.~\ref{fig:benchmark_relative_gains}). IXI reached a repeat mean Dice of 0.9895, which is about 0.0015 above the selected FedCompass calibration target. Camelyon16 reached a repeat mean ROC AUC of 0.7494, about 0.034 above the selected FENS calibration target. Heart Disease matched its selected external target within rounding. TCGA-BRCA found a strong single best C-index during search, but the repeated configuration regressed toward the baseline and should not be treated as a robust gain. ISIC2019 improved substantially over its \nvflare{} baseline but did not reach the strongest selected external target. Descriptive paired mean-difference intervals do not include zero for Heart Disease (+0.0735 $\pm$ 0.0033), IXI (+0.1981 $\pm$ 0.0004), ISIC2019 (+0.1462 $\pm$ 0.0196), and Camelyon16 (+0.1634 $\pm$ 0.0424), but not for TCGA-BRCA (+0.0009 $\pm$ 0.0217). This supports the interpretation that TCGA-BRCA is a seed-sensitive search win rather than a repeated improvement.

\begin{table*}[t]
\centering
\caption{FLamby campaign results.}
\label{tab:flamby}
\resizebox{\textwidth}{!}{%
\begin{tabular}{llllrrrrll}
\toprule
Task & Metric & Baseline repeat & \autoflshort{} repeat & Abs. gain & Rel. gain & External target & Repeat minus target & Target source & Interpretation \\
\midrule
Heart Disease & Accuracy & $0.721 \pm 0.001$ & $0.794 \pm 0.005$ & \scoregain{+0.074} & 10.2\% & 0.794 & +0.000 & FENS iterative FedAvg~\cite{allouah2024fens} & Matches target within rounding \\
TCGA-BRCA & C-index & $0.807 \pm 0.009$ & $0.808 \pm 0.025$ & +0.001 & 0.1\% & 0.815 & -0.007 & FLamby FedAdam mean~\cite{duterrail2022flamby} & Seed-sensitive, no robust lift \\
IXI & Dice & $0.7914 \pm 0.0005$ & $0.9895 \pm 0.0000$ & \scoregain{+0.1981} & 25.0\% & 0.9880 & +0.0015 & FedCompass~\cite{li2024fedcompass} & Mean above target \\
ISIC2019 & Balanced acc. & $0.494 \pm 0.032$ & $0.640 \pm 0.023$ & \scoregain{+0.146} & 29.6\% & 0.750 & -0.110 & FENS iterative FedProx~\cite{allouah2024fens} & Improved, below target \\
Camelyon16 & ROC AUC & $0.5861 \pm 0.0310$ & $0.7494 \pm 0.0182$ & \scoregain{+0.1634} & 27.9\% & 0.7150 & +0.0344 & FENS Fed-Camelyon16~\cite{allouah2024fens} & Mean above target \\
\bottomrule
\end{tabular}}
\end{table*}

\paragraph{IXI check}
Because the IXI gain was unusually large, we verified that the selected
configuration preserved the evaluation path: both baseline and selected
repeats used the same clients, communication rounds, update contract,
data root, and all-client final global-model Dice evaluation. The gain was
not explained by scalar HPO alone. The selected candidate replaced the
small stock FLamby U-Net with a registered residual U-Net-family model
under the fixed 25M-parameter cap, increased local training, and retuned
AdamW regularization and weighted aggregation. Intermediate candidates
improved first with width/residual capacity and then with local-update and
optimizer retuning. Across seeds 42--46, selected scores were highly stable
(0.989--0.990), while matched baselines remained near 0.791. We
therefore interpret IXI as primarily an architecture-capacity and
local-update-budget win under the fixed FL contract.

The kept high-scoring \flamby{} mechanisms varied by task.
Heart Disease benefited from a registered quadratic-linear tabular model and longer local optimization.
IXI improved through a LeakyReLU U-Net variant, local-step adjustments, AdamW-style regularization, and
weighted aggregation.
The main Camelyon16 campaign selected a DSMIL-inspired multiple-instance model for slide classification
\cite{li2021dsmil}, but we treat that mechanism as a hypothesis from the search trace rather than a proven
causal source of the gain.
A completed no-literature, fixed-architecture repeat of a control candidate reached $0.794 \pm 0.024$ ROC AUC
over five seeds, above the literature-enabled campaign repeat mean of $0.749 \pm 0.018$.
A separate repeat of the single-seed sweep winner averaged $0.738 \pm 0.084$, showing that its 0.834 seed-42
score was not a stable repeated result.
Thus, for Camelyon16, the follow-up evidence supports a fixed-architecture recipe-search explanation and
caution about seed sensitivity rather than a causal DSMIL/literature-loop explanation.
ISIC2019 primarily improved through regularization and FedProx-style stabilization, consistent with
overfitting pressure in class-imbalanced dermoscopy.
TCGA-BRCA benefited most from local FedAdam-style interpolation and a reduced server learning rate in the
campaign, but the repeat-seed results caution against treating that single run as a robust target claim.

\subsection{Grouped-Client LEAF Task Profiles}

\begin{figure*}[htbp]
\centering
\begin{tikzpicture}[x=0.42cm,y=0.38cm,font=\scriptsize]
\draw[gray!45] (-1.0,1.30) -- (32.5,1.30);
\foreach \x/\lab in {0/0,10/10,20/20,30/30} {
  \draw[gray!30] (\x,1.25) -- (\x,10.55);
  \node[below,font=\tiny] at (\x,1.25) {\lab\%};
}
\draw[black!60,densely dashed] (0,1.25) -- (0,10.55);
\node[anchor=west,font=\bfseries\scriptsize] at (0,10.78) {\flamby{} healthcare tasks};

\node[anchor=east] at (-1.15,9.92) {ISIC2019};
\fill[nvgreen] (0,9.72) rectangle (29.6,10.12);
\node[anchor=west] at (30.0,9.92) {\scoregain{+29.6\%}};

\node[anchor=east] at (-1.15,9.25) {Camelyon16};
\fill[nvgreen] (0,9.05) rectangle (27.9,9.45);
\node[anchor=west] at (28.3,9.25) {\scoregain{+27.9\%}};

\node[anchor=east] at (-1.15,8.58) {IXI};
\fill[nvgreen] (0,8.38) rectangle (25.0,8.78);
\node[anchor=west] at (25.4,8.58) {\scoregain{+25.0\%}};

\node[anchor=east] at (-1.15,7.91) {Heart Disease};
\fill[nvgreen] (0,7.71) rectangle (10.2,8.11);
\node[anchor=west] at (10.6,7.91) {\scoregain{+10.2\%}};

\node[anchor=east] at (-1.15,7.24) {TCGA-BRCA};
\fill[gray!55] (0,7.04) rectangle (0.1,7.44);
\node[anchor=west] at (0.55,7.24) {+0.1\%};

\draw[black!55] (-1.0,6.35) -- (32.5,6.35);
\node[anchor=west,font=\bfseries\scriptsize] at (0,6.02) {\leaf{} grouped-client profiles};

\node[anchor=east] at (-1.15,5.35) {Shakespeare};
\fill[nvgreen] (0,5.15) rectangle (24.4,5.55);
\node[anchor=west] at (24.8,5.35) {\scoregain{+24.4\%}};

\node[anchor=east] at (-1.15,4.68) {Sent140};
\fill[nvgreen] (0,4.48) rectangle (16.0,4.88);
\node[anchor=west] at (16.4,4.68) {\scoregain{+16.0\%}};

\node[anchor=east] at (-1.15,4.01) {FEMNIST};
\fill[nvgreen] (0,3.81) rectangle (4.6,4.21);
\node[anchor=west] at (5.0,4.01) {\scoregain{+4.6\%}};

\node[anchor=east] at (-1.15,3.34) {Synthetic};
\fill[nvgreen] (0,3.14) rectangle (3.5,3.54);
\node[anchor=west] at (3.9,3.34) {\scoregain{+3.5\%}};

\node[anchor=east] at (-1.15,2.67) {Reddit};
\fill[nvgreen] (0,2.47) rectangle (2.9,2.87);
\node[anchor=west] at (3.3,2.67) {\scoregain{+2.9\%}};

\node[anchor=east] at (-1.15,2.00) {CelebA};
\fill[gray!55] (-0.1,1.80) rectangle (0,2.20);
\node[anchor=west] at (0.55,2.00) {-0.1\%};

\node[below,font=\scriptsize] at (16.0,0.78) {Relative gain over matched repeated baseline};
\end{tikzpicture}
\caption{Mean relative gains over matched repeated baselines across the two benchmark suites (five-seed repeat).
}
\label{fig:benchmark_relative_gains}
\end{figure*}
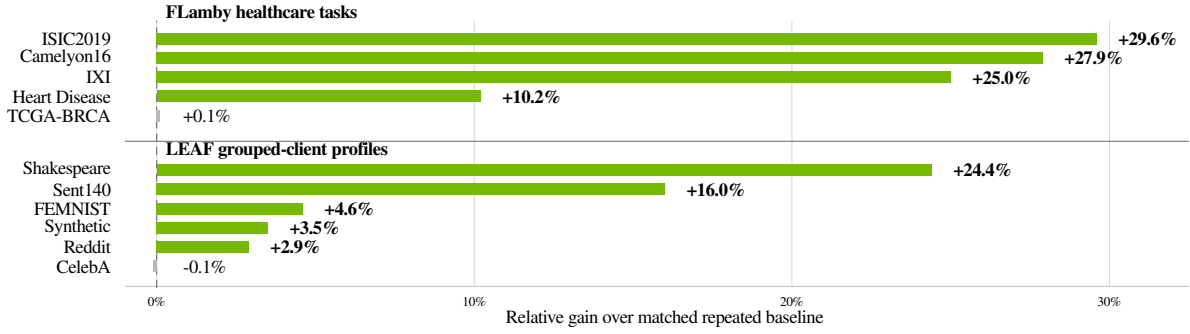

Table~\ref{tab:leaf} summarizes five-seed repeats of the selected grouped-client \leaf{} task-profile
candidates and matched baselines while Fig.~\ref{fig:benchmark_relative_gains} illustrates the gains achieved
across this benchmark.
The repeated results strengthen the evidence on FEMNIST, Sent140, Shakespeare, Synthetic, and Reddit,
where the selected \autoflshort{} candidate remains above the matched baseline mean.
They also expose a search-selected failure case: the CelebA winner from the campaign does not beat the
baseline mean on repeated evaluation; a follow-up top-k repeat check found only a small alternate gain,
whose paired differences still crossed zero.
This result shows why campaign winners should be rerun before they are treated as findings.

\begin{table*}[t]
\centering
\caption{LEAF grouped-client approximation repeat results. 
}
\label{tab:leaf}
\scriptsize
\begin{tabular}{llllrrl}
\toprule
Task & Metric & Baseline repeat & \autoflshort{} repeat & Abs. gain & Rel. gain & Interpretation \\
\midrule
FEMNIST & Accuracy & $0.834 \pm 0.002$ & $0.873 \pm 0.004$ & \scoregain{+0.038} & 4.6\% & Robust repeated gain \\
Shakespeare & Next-char acc. & $0.462 \pm 0.004$ & $0.575 \pm 0.001$ & \scoregain{+0.113} & 24.4\% & Robust repeated gain \\
Synthetic & Accuracy & $0.955 \pm 0.001$ & $0.989 \pm 0.001$ & \scoregain{+0.033} & 3.5\% & Robust repeated gain \\
Sent140 & Accuracy & $0.646 \pm 0.001$ & $0.749 \pm 0.003$ & \scoregain{+0.103} & 16.0\% & Robust repeated gain \\
CelebA & Accuracy & $0.925 \pm 0.004$ & $0.924 \pm 0.006$ & -0.001 & -0.1\% & Baseline-saturated; selected gain did not survive repeat \\
Reddit & Next-token acc. & $0.152 \pm 0.001$ & $0.156 \pm 0.001$ & \scoregain{+0.004} & 2.9\% & Small repeated gain \\
\bottomrule
\end{tabular}
\end{table*}

The same descriptive paired intervals do not include zero for FEMNIST (+0.0383 $\pm$ 0.0050),
Sent140 (+0.1032 $\pm$ 0.0034), Shakespeare (+0.1126 $\pm$ 0.0032),
Synthetic (+0.0333 $\pm$ 0.0017), and Reddit (+0.0044 $\pm$ 0.0009).
CelebA does not separate from zero (-0.0013 $\pm$ 0.0083), reinforcing that selected single-run wins need
repeated evaluation before they become paper claims.

\subsection{FEMNIST Ablation}

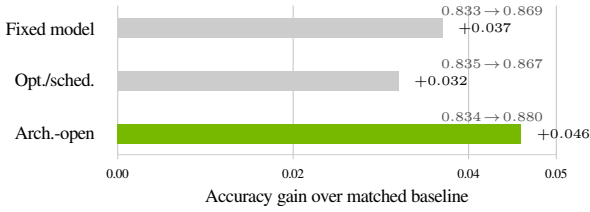
\begin{figure}[htbp]
\centering
\begin{tikzpicture}[x=1cm,y=1cm,font=\scriptsize]
\draw[gray!35] (0,-0.12) -- (5.8,-0.12);
\foreach \x/\lab in {0/0.00,2.32/0.02,4.64/0.04,5.8/0.05} {
  \draw[gray!45] (\x,-0.18) -- (\x,1.85);
  \node[below,font=\tiny] at (\x,-0.18) {\lab};
}

\node[anchor=east] at (-0.18,1.55) {Fixed model};
\fill[gray!40] (0,1.42) rectangle (4.292,1.68);
\node[anchor=west,font=\tiny] at (4.38,1.55) {$+0.037$};
\node[anchor=east,font=\tiny,gray!65!black] at (5.75,1.78) {$0.833 \rightarrow 0.869$};

\node[anchor=east] at (-0.18,0.85) {Opt./sched.};
\fill[gray!40] (0,0.72) rectangle (3.712,0.98);
\node[anchor=west,font=\tiny] at (3.80,0.85) {$+0.032$};
\node[anchor=east,font=\tiny,gray!65!black] at (5.75,1.08) {$0.835 \rightarrow 0.867$};

\node[anchor=east] at (-0.18,0.15) {Arch.-open};
\fill[nvgreen] (0,0.02) rectangle (5.336,0.28);
\node[anchor=west,font=\tiny] at (5.42,0.15) {\scoregain{$+0.046$}};
\node[anchor=east,font=\tiny,gray!65!black] at (5.75,0.38) {$0.834 \rightarrow 0.880$};

\node[below,font=\scriptsize] at (2.9,-0.46) {Accuracy gain over matched baseline};
\end{tikzpicture}
\caption{FEMNIST ablation gains over matched baselines.
}
\label{fig:femnist_ablation}
\end{figure}

To separate the effect of architecture search from fixed-model tuning, we ran three FEMNIST campaign variants (Fig.~\ref{fig:femnist_ablation}). Fixed-model hyperparameter search and optimizer/scheduler search both improved the baseline by roughly 0.03--0.04 accuracy. Allowing registered architecture variants produced the best result, improving the baseline by 0.046. This supports the claim that code-level mutations can add value beyond what a traditional scalar hyperparameter sweep can.


\subsection{Cross-Task Patterns}

Several patterns recur across tasks.
First, FedProx-like proximal regularization was frequently useful on heterogeneous and noisy tasks,
including ISIC2019, FEMNIST, Shakespeare, and Sent140; on CelebA, it produced the largest selected score but
did not survive repeat evaluation.
Second, server-side momentum or FedOpt-style scaling helped when the default aggregated model difference was
too conservative for the fixed round budget.
Third, task-specific architecture variants mattered most when the baseline architecture was clearly
underpowered or mismatched to the data representation, as in IXI, Heart Disease, and FEMNIST.
Camelyon16 is a useful counterexample: the campaign trace suggested an architecture mechanism, but a
no-literature fixed-architecture repeat later exceeded the literature-enabled repeat mean.
Finally, discarded candidate rows made plateaus visible and encouraged the agent to switch from parameter
jitter to source-backed proposals.

The logged literature loop had mixed effects.
Among the main \flamby{} campaigns with literature events, the best post-literature score exceeded the best
pre-literature score for TCGA-BRCA (0.8426 vs. 0.8411), IXI (0.9896 vs. 0.9894), and Camelyon16
(0.7780 vs. 0.5478, where the early literature loop preceded the DSMIL-style MIL search).
However, the completed no-literature/fixed-architecture Camelyon16 sweep reached a single-seed best of
0.8344, and a five-seed repeat of a fixed-architecture control candidate reached $0.794 \pm 0.024$,
exceeding the literature-enabled repeat mean.
Repeating the single-seed sweep winner gave $0.738 \pm 0.084$, below the literature-enabled repeat mean and
demonstrating why selected single-seed winners should not be promoted without repeats.
It did not improve the final best score for Heart Disease or ISIC2019, and the Sent140 \leaf{} literature
events also did not exceed the earlier best.
A 101-row Sent140 no-literature local-sweep ablation reached 0.7545, essentially matching the main Sent140
campaign best under the same grouped-client harness.
This suggests that literature-grounded recovery can generate useful candidates, but the available ablations
do not show that the final gains depended causally on literature-derived proposals.

\subsection{Mechanism Attribution and Controls}

\begin{table*}[t]
\centering
\caption{Condensed FL-mechanism attribution for selected \autoflshort{} improvements.
Gains are relative to matched repeated baselines unless noted.}
\label{tab:mechanisms}
\small
\setlength{\tabcolsep}{4pt}
\renewcommand{\arraystretch}{1.16}
\begin{tabular}{@{}p{0.14\textwidth}p{0.14\textwidth}p{0.43\textwidth}p{0.22\textwidth}@{}}
\toprule
Task & Repeat effect & Repeat-supported mechanism & FL interpretation \\
\midrule
IXI &
\scoregain{+0.198 Dice} &
Wider residual U-Net-family client model under the 25M cap, longer local training, and weighted
FedAvg-style aggregation~\cite{cicek20163dunet,mcmahan2017fedavg}. &
Federated model-capacity and local-update-budget win. \\
\addlinespace[0.35em]
Heart Disease &
\scoregain{+0.074 acc.} &
Quadratic-linear client model with coordinate-wise median-style robust server aggregation and 200 local
steps~\cite{yin2018byzantine}. &
Robust aggregation plus task-local FL recipe win; rediscovered in two of three independent trajectories. \\
\addlinespace[0.35em]
FEMNIST &
\scoregain{+0.038 acc.} &
Registered client CNN variant with FedProx-style local objective and optimizer/scheduler retuning under the
same grouped-client contract~\cite{li2020fedprox}. &
Architecture-open win; fixed-model controls improved, but stayed below the selected architecture-open run. \\
\addlinespace[0.35em]
Sent140 &
\scoregain{+0.103 acc.} &
Longer local updates and FedProx-style local objective on the existing model~\cite{li2020fedprox}. &
Fixed-surface FL recipe tuning; scripted HPO and no-literature sweep reproduced the gain. \\
\addlinespace[0.35em]
ISIC2019 &
\scoregain{+0.146 bal. acc.} &
FedProx-style client-objective stabilization and regularization under the fixed pretrained client
model~\cite{li2020fedprox}. &
Client-objective stabilization under the fixed architecture. \\
\addlinespace[0.35em]
Camelyon16 &
\scoregain{+0.163 ROC AUC} &
Fixed-architecture FL recipe survived repeat checks; DSMIL-style MIL was explored but not supported as the
causal mechanism~\cite{li2021dsmil}. &
FL recipe win; repeat checks revised the architecture attribution. \\
\addlinespace[0.35em]
CelebA &
No repeat gain &
FedProx-style client-objective candidate selected from a single campaign score~\cite{li2020fedprox}. &
Selected-run artifact; five-seed repeat did not beat the baseline. \\
\bottomrule
\end{tabular}
\end{table*}

\paragraph{Mechanism attribution}
Table~\ref{tab:mechanisms} summarizes the main FL mechanisms supported by the controls.
For an FL audience, the strongest results are not the largest raw score changes alone, but the cases where
\autoflshort{} changed the federated recipe while preserving the communication and scoring contract:
server aggregation in Heart Disease, client capacity and local-update budget in IXI, and grouped-client
architecture/update choices in FEMNIST.
Sent140 and ISIC2019 show that strong gains can also arise from careful tuning of existing FL recipe knobs.
Camelyon16 and CelebA show why search-selected mechanisms need repeat checks before being treated as causal
findings.

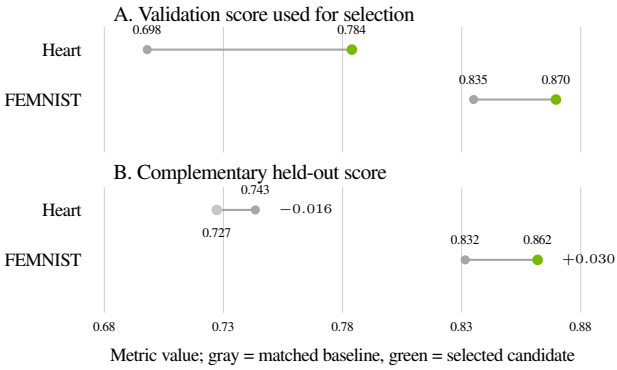
\begin{figure}[htbp]
\centering
\begin{tikzpicture}[x=1cm,y=1cm,font=\scriptsize]
\node[anchor=west,font=\footnotesize] at (0,2.75) {A. Validation score used for selection};
\foreach \x/\lab in {0/0.68,1.575/0.73,3.15/0.78,4.725/0.83,6.3/0.88} {
  \draw[gray!35] (\x,0.92) -- (\x,2.58);
}
\node[anchor=east] at (-0.18,2.28) {Heart};
\draw[gray!70,thick] (0.567,2.28) -- (3.273,2.28);
\fill[gray!70] (0.567,2.28) circle (0.06);
\fill[nvgreen] (3.273,2.28) circle (0.07);
\node[above,font=\tiny] at (0.567,2.36) {0.698};
\node[above,font=\tiny] at (3.273,2.36) {0.784};

\node[anchor=east] at (-0.18,1.62) {FEMNIST};
\draw[gray!70,thick] (4.883,1.62) -- (5.972,1.62);
\fill[gray!70] (4.883,1.62) circle (0.06);
\fill[nvgreen] (5.972,1.62) circle (0.07);
\node[above,font=\tiny] at (4.883,1.70) {0.835};
\node[above,font=\tiny] at (5.972,1.70) {0.870};

\node[anchor=west,font=\footnotesize] at (0,0.63) {B. Complementary held-out score};
\foreach \x/\lab in {0/0.68,1.575/0.73,3.15/0.78,4.725/0.83,6.3/0.88} {
  \draw[gray!35] (\x,-1.20) -- (\x,0.46);
  \node[below,font=\tiny] at (\x,-1.20) {\lab};
}
\node[anchor=east] at (-0.18,0.16) {Heart};
\draw[gray!70,thick] (1.997,0.16) -- (1.487,0.16);
\fill[gray!70] (1.997,0.16) circle (0.06);
\fill[gray!45] (1.487,0.16) circle (0.07);
\node[above,font=\tiny] at (1.997,0.24) {0.743};
\node[below,font=\tiny] at (1.487,0.04) {0.727};
\node[anchor=west,font=\tiny] at (2.18,0.16) {$-0.016$};

\node[anchor=east] at (-0.18,-0.50) {FEMNIST};
\draw[gray!70,thick] (4.772,-0.50) -- (5.730,-0.50);
\fill[gray!70] (4.772,-0.50) circle (0.06);
\fill[nvgreen] (5.730,-0.50) circle (0.07);
\node[above,font=\tiny] at (4.772,-0.42) {0.832};
\node[above,font=\tiny] at (5.730,-0.42) {0.862};
\node[anchor=west,font=\tiny] at (5.93,-0.50) {\scoregain{$+0.030$}};

\node[below,font=\scriptsize] at (3.15,-1.54) {Metric value; gray = matched baseline, green = selected candidate};
\end{tikzpicture}
\caption{Validation-selected and held-out-reported check.
}
\label{fig:heldout_check}
\end{figure}

Figure~\ref{fig:heldout_check} reports the held-out check: Heart Disease exposed a validation-selected false
positive, while FEMNIST retained a +0.030 held-out gain.
This reinforces the main evaluation rule: search scores drive campaigns, but repeat and held-out evaluations
decide which findings become scientific claims.

\section{Discussion}

The results show that a coding agent can be productive in FL research when it operates inside a fixed
execution harness.
\autoflshort{} did not merely tune a static list of hyperparameters: it introduced task-specific model
architectures, optimization rules, and regularization strategies while preserving final global-model scoring.
This is important because many useful FL improvements require code changes that do not fit neatly into
conventional HPO.
The causal evidence is strongest when the same mechanism is recovered by
independent trajectories or when the same-budget controls isolate a smaller search
surface; otherwise, we report the mechanism as an attribution supported by the
search trace rather than as a single-factor causal effect.

The same mechanism also introduces limitations.
Agentic searches are not deterministic, and a single campaign best can be seed-sensitive.
TCGA-BRCA illustrates this: the campaign found a strong best run, but the selected configuration did not
reproduce the same margin across follow-up seeds.
CelebA shows the same issue in the \leaf{} block: the selected winner was slightly below the matched baseline
after five seeds, and a targeted top-k repeat check found only a small, uncertain alternate gain.
ISIC2019 also demonstrates that improving a local \nvflare{} baseline does not imply beating a strong
published target, especially when that target may use a different protocol, personalization strategy,
or evaluation budget.
For \leaf{}, our grouped-client adapter is a practical approximation rather than an exact reproduction of the
original large-scale cross-device setting.
These negative and mixed cases motivate the distinction between search scores, repeated results, and held-out
checks.
The results therefore support \autoflshort{} as a tool for generating and triaging FL candidates, not as an
autonomous source of final benchmark claims.
We also do not claim that the observed trajectories are independent of the
chosen coding-agent backend; evaluating multiple agent models under the same
task profiles and budgets is an important next step.

The current implementation also does not solve broader governance questions around autonomous research.
The agent can waste compute, overfit to a benchmark, or cite literature too shallowly if the prompt and review
rules are weak.
The records and fixed mutation surface reduce these risks but do not remove the need for human scientific
review.
This study adds five-seed \flamby{} and \leaf{} repeats, same-budget scripted HPO controls for FEMNIST, Heart Disease, ISIC2019, and Sent140, a Sent140 no-literature ablation, Heart agent-trajectory repeats, and two
validation-selected/held-out-reported checks.
The held-out checks are especially instructive: Heart Disease exposed a validation-selected false positive,
whereas FEMNIST retained a held-out gain.
Stronger causal claims would still benefit from randomized, Bayesian, or
evolutionary HPO/NAS controls under the same candidate budget across more
tasks, additional no-literature controls for architecture-heavy discoveries,
and externally defined validation/test splits.

\section{Conclusion}

We presented Auto-FL-Research, a constrained agentic workflow for FL research on top of \nvflare{}.
\autoflshort{} uses task profiles, explicit mutation surfaces, fixed budgets, immutable FL communication
contracts, final global-model scoring, and persistent candidate records.
This setup lets agents explore code-level FL recipe changes while preserving comparability across candidates.
Across healthcare \flamby{} tasks and grouped-client LEAF profiles, \autoflshort{} selected higher-scoring
candidates under the fixed harness.
Five-seed repeats supported gains on four of five \flamby{} tasks and five of six \leaf{} profiles,
while TCGA-BRCA and CelebA showed why selected wins require repeat evaluation.
Same-budget controls further showed that the most informative wins came from FL mechanisms such as robust
aggregation, local-update budgeting, client-objective stabilization, and architecture choices under a fixed
communication contract.
Other gains were recovered by fixed-surface tuning or failed under repeat or held-out evaluation.
The main result is therefore not an unconstrained benchmark claim; it is a practical workflow for generating,
recording, and checking FL candidate algorithms.

\section*{Code Availability and AI Use Disclosure}
The code artifact corresponding to the methods described in this paper is
available as the NVIDIA FLARE Auto-FL research example: \url{https://github.com/NVIDIA/NVFlare/tree/main/research/auto-fl-research}.
Codex was used to help set up and run agentic experiments, and
Codex and ChatGPT assisted with initial drafting and editing. The authors
reviewed, verified, and approved all results, claims, citations, figures,
tables, and final text, and remain responsible for the paper.

\small
\bibliographystyle{myIEEEtran}
\bibliography{references}

\end{document}